\documentclass{article}

\usepackage{amsmath}
\usepackage{graphicx}

\usepackage[preprint]{nips_2018}

\usepackage[utf8]{inputenc} % allow utf-8 input
\usepackage[T1]{fontenc}    % use 8-bit T1 fonts
\usepackage{hyperref}       % hyperlinks
\usepackage{url}            % simple URL typesetting
\usepackage{booktabs}       % professional-quality tables
\usepackage{amsfonts}       % blackboard math symbols
\usepackage{nicefrac}       % compact symbols for 1/2, etc.
\usepackage{microtype}      % microtypography

\title{Optimal Algorithms for Ski Rental with Soft Machine-Learned Predictions}

\author{
  Rohan Kodialam \\
  Massachusetts Institute of Technology\\
  Cambridge MA \\
  \texttt{kodialam@mit.edu} \\
}

\begin{document}

\maketitle

\begin{abstract}
  We consider a variant of the classic Ski Rental online algorithm with applications to machine learning. In our variant, we allow the skier access to a black-box machine learning algorithm that provides an estimate of the probability that there will be less than a threshold number of ski-days. We derive a class of optimal randomized algorithms to determine the strategy that minimizes the worst-case expected competitive ratio for the skier given a prediction from the machine learning algorithm, and analyze the performance and robustness of these algorithms. 
\end{abstract}

\section{Introduction}
Online decision-making problems fundamentally address the issue of dealing with the uncertainty inherently present in the future. In broad terms, these problems can be addressed in two ways. First, a predictive approach like a machine learning algorithm can be used to guess at future events and to act accordingly. This method, while clearly powerful, has the drawback that it is very difficult to make any guarantees about the performance of the algorithm. 

Another paradigm for solving these problems is to use competitive analysis to guarantee a bound on the performance of a given algorithm. In this approach, we consider some cost function and note the cost that would be incurred by an omniscient algorithm that could access future data. Then, we compare this optimal omniscient cost to the worst-case cost of an online algorithm which cannot use future data. By bounding the ratio of these two costs, a guarantee can be made as to how well the online algorithm will perform in terms of the performance of an omniscient algorithm.

A classic problem in competitive analysis is the \emph{Ski Rental Problem}. In this problem, a skier must decide whether to rent skis at a rate of $\$ 1$ per day, or to buy skis at a price of $\$ B$. The uncertainty lies in the number of ski days left in the season - if there are very few days, it is optimal to rent the skis. On the other hand, if there are many ski days it may be more cost-effective to buy the skis outright. This simple paradigm extends to various practical problems, such as a firm deciding between purchasing servers or using an on-demand cloud computing solution to provide the computational power needed to satisfy an unknown future demand. 

\subsection{Prior Work \& Our Contribution}
Recent work has shown promising results in combining the predictive and competitive-analysis paradigms for approaching online problems, with applications to various fields  [1,2,5,8]. In this paper, we consider the problem of using predictions from a machine learning algorithm to improve the Ski Rental online algorithm in a novel way. Past work has dealt with the case of a direct prediction of the number of ski days and using this prediction to make a binary decision on whether there will be more or less than $B$ ski days [1]. This work also establishes a useful model in which the combination of a confidence parameter $\lambda$ and a prediction of whether there will be more or less than $B$ ski days is used to construct a strategy with guarantees on the expected competitive ratio both when the prediction is correct and when it is wrong - this paradigm effectively captures the need to measure the robustness of any algorithm that uses potentially erroneous data provided by a learning algorithm. 

We note that many machine learning algorithms such as neural nets make hard predictions (i.e. there are at most $B$ ski days) by thresholding a soft prediction. That is, in order to predict if event $E$ will occur, a typical machine learning algorithm will estimate the probability that $E$ occurs, and predict that $E$ will occur if this probability is greater than $\nicefrac{1}{2}$. In the context of the Ski Rental problem, predicting if there will be more or less than $B$ ski days then necessitates predicting the probability that there will be at most than $B$ ski days. 

In this paper, we present a more general result in which we allow the skier to use a black-box machine learning algorithm to get a prediction on the probability that there will be at most $B$ ski days. We then find the optimal randomized algorithm that the skier should use to minimize the expected worst-case competitive ratio as compared to an omniscient optimal algorithm. In section 2.1 we explicitly find the expected competitive ratio attained by our machine learning augmented algorithm. Furthermore in section 2.3, we address the issue of robustness and explicitly find worst-case bounds on the expected competitive ratio of our algorithm when it uses erroneous predictions. 

\subsection{The Ski Rental Problem}

The Ski Rental problem is often stated as a discrete problem - for ease of computation we define and use the following continuous analogue as our model. We consider a skier  who chooses at each moment of time $t$ whether to rent skis at a rate of $1$ dollar per unit time, or to buy the skis for $B$ dollars. We denote the moment the skier buys the skis by $x$.

The skier's challenge is that the amount of time $y$ that skiing is possible is unknown. In the offline case, we allow the skier access to $y$. By inspection, the optimal solution is then to rent until $y$ if $y<B$ and to buy immediately (i.e. pick $x=0$) otherwise - this strategy will incur a cost of $OPT(y) = \min(y,B)$.

To measure the performance of an online algorithm, we use the metric of the worst-case expected competitive ratio, which we can informally define as the expected ratio between the worst-case cost of our online algorithm and the cost incurred by an optimal omniscient algorithm that knows the future. While this is not the only metric possible and other metrics like the average-case competitive ratio of [7] may give more optimistic results, this metric allows us to account for worst case behaviour and is thus very robust. We now formally define the worst-case competitive ratio.

\subsection{The Competitive Ratio}
To get a worst-case measure of performance, we introduce an adversary who chooses $y$ in order to maximize the skier's cost. We note that both the skier and the adversary may use probabilistic strategies to decide on $x$ and $y$ respectively. Thus, the spaces of strategies for the skier and the adversary are $$\mathcal{P} = \left\{p(x) \text{ such that } p(x)\geq 0 \ \forall \ x, \int_0^\infty p(x)dx = 1\right\}$$ and $$\mathcal{Q} = \left\{q(y) \text{ such that } q(y)\geq 0 \ \forall \ y, \int_0^\infty q(y)dy = 1\right\}$$ respectively. Let $C(x,y)$ denote the cost incurred by the skier when the skier plays pure strategy $x$ and the adversary plays pure strategy $y$ - this cost will be
\begin{equation}
C(x,y) = \begin{cases} 
      x+B & x < y \\
      y & x \geq y
   \end{cases}
\end{equation}
and as before the optimal offline cost that could have been incurred will be 
$$ OPT(y) = \begin{cases} 
      B & y \geq B \\
      y & y < B
   \end{cases}.
$$
We measure the performance of an algorithm by the \emph{competitive ratio} defined by the ratio of the cost incurred by the algorithm to the optimal offline cost, thus the competitive ratio for pure strategies $x$ and $y$ is given by 
\begin{equation}
    CR(x,y) = \frac{C(x,y)}{OPT(y)}.
\end{equation}

The metric of interest for a strategy $p$ is how well it fares in expectation against an optimal adversary. Thus, for any strategy $p$ we define the \emph{worst case expected competitive ratio} by 
$$ J(p) = \max_{q \in \mathcal{Q}} \iint CR(x,y) p(x) q(y) dx dy.$$
\subsection{Reducing the Adversary's Strategy Space}
In the previous section, we allowed the adversary to choose any strategy in the space $\mathcal{Q}$, which is the space of all valid probability distributions with support $[0,\infty)$. We now perform a brief game-theoretic analysis to find a smaller space of strategies that the adversary will always choose from. In doing so, we interpret the Competitive Ratio as the payoff of a zero sum game played by the adversary and the skier as in [6]. Then, using techniques similar to those in [9], we can find pure strategies for the adversary that are strictly dominated by other pure strategies, and thereby eliminate these dominated pure strategies from the adversary's strategy space. 

Consider $y \geq B$, and some $y'>y$. Then, 
$$ CR(x,y) = \frac{1}{B} \begin{cases} 
      x+B & x < y \\
      y & x \geq y
   \end{cases}
$$
and 
$$ CR(x,y') = \frac{1}{B} \begin{cases} 
      x+B & x < y' \\
      y' & x \geq y'
   \end{cases}
   = \frac{1}{B} \begin{cases} 
      x+B & x < y \\
      x+B & y \leq x < y' \\
      y' & x \geq y'
   \end{cases}
$$
We can thus observe that when $x\in(0,y)$, $CR(x,y') = CR(x,y)$. When $x\in[y,y')$, $CR(x,y') = \frac{x}{B} + 1 \geq \frac{y}{B} + 1 > \frac{y}{B} = CR(x,y)$. Finally, for $x\in[y',\infty)$, $CR(x,y') = \frac{y'}{B}  > \frac{y}{B} = CR(x,y)$.

Thus, for all $x$, $CR(x,y') \geq CR(x,y)$ and for some $x$ (namely, those in $[y,\infty)$), $CR(x,y') > CR(x,y)$. We thus conclude that for the adversary, any strategy $y \geq B$ is dominated by a strategy $y'>y$ and thus we can reduce the space of choices of $y$ from distributions with support $[0,\infty)$ to those with support $[0,B) \cup \{\infty\}$, where $y=\infty$ represents the case where there are infinite ski days.

\section{Finding the Optimal Strategy with Information}

Our model of the Ski Rental problem as a zero-sum game has the added benefit that constraints on the distribution of ski days can be interpreted as constraints on the strategy space $\mathcal{Q}$ of the adversary. As explained before, we constrain the probability of the number of ski days being at most $B$. Let this probability be $\alpha$ - then, the strategy space for the adversary becomes 
$$\mathcal{Q}_\alpha = \left\{q(y) \text{ such that } q(y)\geq 0 \ \forall \ y\in[0,B], \int_0^B q(y)dy = \alpha\right\}.$$ That is, the adversary can choose a distribution with total probability $\alpha$ between $0$ and $B$, and must allocate the remaining $1-\alpha$ probability to the case of $y=\infty$.

To find the best strategy for the skier, we seek to find $p\in \mathcal{P}$ that is the minimizer over all $p \in \mathcal{P}$ of  
$$ J(p;\alpha)  =  \max_{q \in \mathcal{Q}_\alpha} \iint \frac{C(x,y)}{OPT(y)} p(x) q(y) dx dy.$$
To do so, we first fix an arbitrary value of $y\in[0,B]$. Then, let $\mathcal{C}(p,y)$ be the expected cost incurred if there are $y$ ski days and if the day $x$ on which skis are bought is distributed as $p$:

$$ \mathcal{C}(p,y) = \int_0^y (x+B) p(x) dx + \int_y^\infty y p(x) dx$$

The expected competitive ratio will therefore be 

$$ \int_0^B \frac{\mathcal{C}(p,y)}{y} q(y) dy + q(\infty)\int_0^\infty \frac{x+B}{B} p(x) dx$$

We now use the constraint that $\int_0^B q(y) dy = \alpha$ and associate a Lagrange multiplier $\lambda$ with this constraint. Since $q$ must be a normalized PDF, this implies that $q(\infty) = 1-\alpha$. Thus, the adversary's problem becomes
$$ \max_q \int_0^B (\frac{\mathcal{C}(p,y)}{y} - \lambda) q(y) dy + (1-\alpha)\int_0^\infty \frac{x+B}{B} p(x) dx + \alpha \lambda$$

The Lagrange dual to this problem will then be to minimize 
$$ L = (1-\alpha)\int_0^\infty \frac{x+B}{B} p(x) dx + \alpha \lambda$$
over all $p\in \mathcal{P}$ and $\lambda$, with the constraint that $\frac{\mathcal{C}(p,y)}{y} \leq \lambda$ for all $y \in [0,B]$. Consider the case where the constraint is tight. This implies that $\mathcal{C}(p,y) = \lambda y$.

We note that this result should hold for all $y\in[0,B]$, so we can take two derivatives of this relation with respect to $y$ to find that $\partial^2_y \mathcal{C}(p,y) = 0$. This gives the differential equation $B\partial_x p(x) = p(x)$, which implies $p(x) = K e^{\frac{x}{B}}$ for some constant $K$.

We note that in order for this $p(x)$ to be a PDF, it must have support only in some interval $[0,a)$. Then, we can write the distribution as 
\begin{equation}
 p(x) = \begin{cases} 
      \frac{e^{\frac{x}{B}}}{B \left(e^{a/B}-1\right)} & \text{if } 0\leq x < a \\
      0 & \text{otherwise}
   \end{cases}
\end{equation}

We make the following observations about limiting cases of the problem: First, as $\alpha$ goes to $0$, it is certain that there will be more than $B$ snow days, and thus the optimal strategy is to buy skis immediately. This would mean $a \to 0$. On the other hand, if $\alpha$ goes to $1$ there will be less than $B$ snow days, so the optimal strategy is to never buy. This effectively means that the designer will choose to buy on a very large day, and as such $a$ should go to $\infty$. 

The constraint $\frac{\mathcal{C}(p,y)}{y} \leq \lambda$ for all $y \in [0,B]$ will be tight if $a\geq B$. On the other hand, the constraint is not tight for $a<B$ and thus we must explicitly find $\lambda = \max_{y\in[0,B]} \frac{\mathcal{C}(p,y)}{y}$.

Using this truncated PDF, we find that 
$$ C(p,y) = \left(\frac{1}{e^{a/B}-1}+1\right) \begin{cases} 
      y & \text{if } y < a \\
      a & \text{if } y \geq a
   \end{cases}
$$
and thus the maximum value $\max_{y\in[0,B]} \frac{\mathcal{C}(p,y)}{y} = \left(\frac{1}{e^{a/B}-1}+1\right)$.

We also find that 
$$ \int_0^\infty \frac{x+B}{B} p(x) dx  = \frac{a}{B} \left(\frac{1}{e^{a/B}-1}+1\right)$$
and adding the terms together the value of the dual objective function can be expressed in terms of only the cutoff value $a$ as 

$$L(a) = \frac{e^{a/B} (a + \alpha  (B-a))}{B \left(e^{a/B}-1\right)}.$$ 

For ease of notation, we let $z = \frac{a}{b}$. Then, we can rewrite the objective function value as 

\begin{equation}
    L(z)=\frac{e^{z} (z + \alpha(1-z))}{\left(e^{z}-1\right)}.
\end{equation}

The dual objective is minimized by varying $z$, and taking a derivative with respect to $z$ gives 

$$\partial_z L(z) = -\frac{e^z \left(-\alpha  z+(\alpha -1) e^z+z+1\right)}{\left(e^z-1\right)^2}.$$

Thus, the minimizing $z$ will solve $(\alpha-1)e^z - (\alpha-1)z + 1 = 0 $. While this is not solvable in terms of elementary functions, we note that it will always have a non-negative root so long as $1/(1-\alpha)>1$, which will be the case for any valid $\alpha\in[0,1]$. Thus, for a given $\alpha$, we can find a value $z^*(\alpha)$ such that the value of the dual objective function is minimized. By strong duality, this same choice of $z^*$ will solve the primal and thus give the distribution $p(x)$ with the optimal worst-case expected competitive ratio.

We can further confirm this result by noting in the limiting case $\alpha=0$, this equation reduces to $e^z - z - 1 =0$, which is clearly solved by $z^*(0)=0$, giving a cutoff of $a = Bz = 0$. Likewise, as $\alpha \to 1$, we seek to solve $e^z - z = \frac{1}{1-\alpha}$, which will be solved by $z^*(1) \to \infty$ as $\alpha\to 1$ since $e^z-z$ is an increasing function for positive $z$. 

\subsection{Analysis of the Optimal Cutoff}

We have found that an optimal skier will choose to buy skis at day $x$ by drawing a value from the distribution $$ p(x) = \begin{cases} 
      \frac{e^{\frac{x}{B}}}{B \left(e^{z^*(\alpha)}-1\right)} & \text{if } 0\leq x < Bz^*(\alpha) \\
      0 & \text{otherwise}
   \end{cases}
$$
where $z^*(\alpha)$ solves the equation $(\alpha-1)e^z - (\alpha-1)z + 1 = 0 $. The solution to this equation can be expressed in terms of the negative branch of the Lambert $W$ function\footnote{Recall that the branches of the Lambert $W$ function satisfy the equation $W(x)^{W(x)}=x$} as 
\begin{equation}
z^*(\alpha) =\frac{1}{\alpha-1}-W_{-1}(-e^{\frac{1}{\alpha-1}}).
\end{equation}
The behaviour of this function can be seen in Figure 1.
Furthermore, plugging this value of $z$ back into the expression for $p(x)$, and then using $p(x)$ to compute the optimal value of the objective function, i.e. the worst case expected competitive ratio, we find that as we vary $\alpha$, 
\begin{equation}
\min_{p\in \mathcal{P}}J(p;\alpha) = (\alpha - 1) W_{-1}(-e^{\frac{1}{\alpha-1}}),
\end{equation}
as can be seen in Figure 2.

Looking at figure 2, it is interesting to note that the highest (i.e. worst) competitive ratio occurs at a value of $\alpha$ less than $\nicefrac{1}{2}$. Specifically, we can maximize the worst case expected competitive ratio from Equation 6 over all $\alpha \in [0,1]$ to find that the maximizing $\alpha$ is $\frac{e-2}{e-1} \approx 0.42$ and that the maximum competitive ratio is $\frac{e}{e-1} \approx 1.58$. As shown in previous works [3], the competitive ratio of $\frac{e}{e-1}$ is the best possible competitive ratio attainable by any randomized algorithm for the ski rental problem without any additional information. Thus, we see that having $\alpha = \frac{e-2}{e-1}$ is the same as having no information, and that in all other cases the competitive ratio decreases from this no-information value. Furthermore, we note that the maximizing $\alpha$ is not at $\nicefrac{1}{2}$, and thus the results shown in works like [1] are not as efficient as possible in the region of $\frac{e-2}{e-1} < \alpha < \frac{1}{2}$, where it is more likely that $y$ is less than $B$, yet the optimal cutoff value for $p(x)$ lies beyond $B$. 
\begin{figure}
\centering
\begin{minipage}{.5\textwidth}
  \centering
  \includegraphics[width=\linewidth]{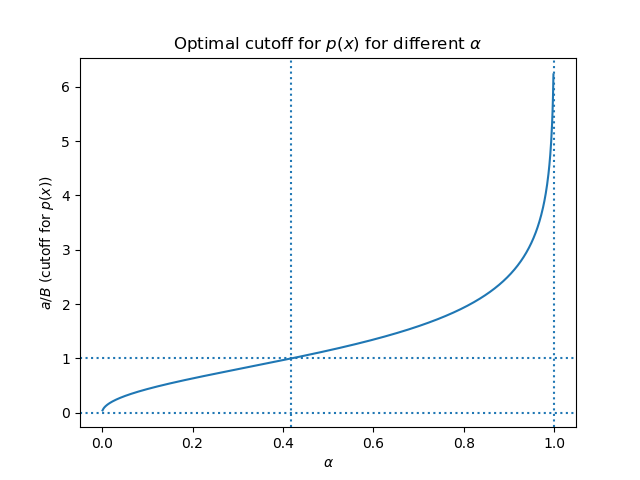}
  \caption{Optimal Cutoff of $p(x)$ for different $\alpha$}
  \label{fig:test1}
\end{minipage}%
\begin{minipage}{.5\textwidth}
  \centering
  \includegraphics[width=\linewidth]{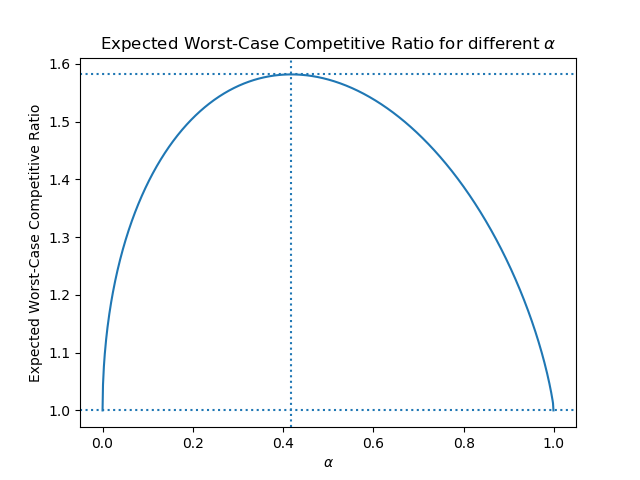}
  \caption{Optimal Expected Worst-Case Competitive Ratio for different $\alpha$}
  \label{fig:test2}
\end{minipage}
\end{figure}

\subsection{The Adversary's Strategy}

We can use a similar analysis to find the strategy that will be used by the adversary. Recall that the adversary is constrained to pick $q(\infty)=1-\alpha$, and is free to pick the remaining $q(y)$ such that $\int_0^B q(y) = \alpha$. The expected competitive ratio if the skier buys on day $x$ and if the adversary picks $y$ according to the distribution $q(y)$ is 
$$\mathcal{C}(x,q) = \int_0^x q(y) dy + \int_x^\infty \frac{x+B}{y} q(y) dy + (1-\alpha)\frac{x+B}{B}$$
We note that the skier is constrained to pick $p(x)$ such that $\int_0^\infty p(x) = 1$, and thus adding in a Lagrange multiplier for this constraint, the relevant terms in the expected competitive ratio will be $$ \int_0^B \left(\frac{\mathcal{C}(x,q)}{y}-\lambda\right) p(x) dx + \lambda$$ and the dual problem will be to minimize $\lambda$ while keeping $\lambda \geq \frac{\mathcal{C}(x,q)}{y}$. As before, taking two derivatives gives $q(y) = K y e^{-y/B}$. Furthermore, we know that this function should be normalized such that $\int_0^B q(y) dy = \alpha$, so $q(y) = \frac{\alpha  y e^{1-\frac{y}{B}}}{(e-2) B^2}$.

\subsection{Sensitivity of the Skier's Strategy to ML Errors}
Our analysis so far has focused on the case where the probability $\alpha$ correctly reported. Most machine learning algorithms, of course, will not always report precise values and thus we now analyze the case where a mistake is made in the prediction of $\alpha$. 

Recall from Equation 4 that if a cutoff $a = zB$ is chosen by the skier, the expected competitive ratio will be $$L(z)=\frac{e^{z} (z + \alpha(1-z))}{\left(e^{z}-1\right)}.$$ Consider the case where an ML algorithm has reported a value $\hat{\alpha}$ for the probability that $y$ is at most $B$, and where the true value of $\alpha$ is between $\hat{\alpha}-\epsilon$ and $\hat{\alpha}+\epsilon$. Then, if the skier picks a cutoff based on the possibly faulty value $\hat{\alpha}$, the expected competitive ratio will lie between
$$\frac{e^{z^*(\hat{\alpha})} (z^*(\hat{\alpha}) + (\hat{\alpha}-\epsilon)(1-z^*(\hat{\alpha})))}{\left(e^{z^*(\hat{\alpha})}-1\right)} \text{ \ and \ } \frac{e^{z^*(\hat{\alpha})} (z^*(\hat{\alpha}) + (\hat{\alpha}+\epsilon)(1-z^*(\hat{\alpha})))}{\left(e^{z^*(\hat{\alpha})}-1\right)}.$$ 

Note that the deviation from the competitive ratio in either case can be expressed by 
$$
\pm \epsilon e^{z^*(\hat{\alpha})} \left(  \frac{1-z^*(\hat{\alpha})}{e^{z^*(\hat{\alpha})}-1} \right)
$$

In the worst case, therefore, the increase in expected competitive ratio per unit increase in the error $\epsilon$ will be 
\begin{equation*}
\Delta(\hat{\alpha}) = \left| (1-z^*(\hat{\alpha}))\frac{e^{z^*(\hat{\alpha})}}{e^{z^*(\hat{\alpha})}-1} \right| = \left| \frac{1}{1-\hat{\alpha}} +W_{-1}(-e^{\frac{1}{\hat{\alpha}-1}}) + \frac{1}{(\hat{\alpha}-1)\left(1+W_{-1}(-e^{\frac{1}{\hat{\alpha}-1}})\right)}\right|.
\end{equation*} 

While this function is quite complicated, it has several key properties. First, at $\hat{\alpha}=0$ and $\hat{\alpha}=1$, $\Delta(\hat{\alpha})$ goes to infinity. Indeed, if the skier is told that it will certainly snow less than $B$ days, they will choose to never buy. However, if there is an error of size $\epsilon$ then with probability $\epsilon$ there will be infinite ski days. Thus, the skier will pay a cost of $\infty$ when the optimal cost would have been $B$, incurring an infinite competitive ratio that will make the overall expected competitive ratio go to infinity as well. Likewise, if the skier is told it will certainly snow more than $B$ days and there is an error, the skier will face an infinite competitive ratio if it snows zero days since buying incurs a cost of $B$ yet the optimal scheme would incur a cost of $0$.

Another key detail is that $\Delta(\hat{\alpha}) = 0$ when $\hat{\alpha} = \frac{e-2}{e-1}$. This means that even if there is error in the probability estimate, the competitive ratio will remain $\frac{e}{e-1}$. This lines up with the idea that $\hat{\alpha} = \frac{e-2}{e-1}$ actually does not give any information to the skier, who therefore picks the optimal strategy for the ski rental problem without any additional information. Thus, since the skier is agnostic to $\hat{\alpha}$, errors in this value will not worsen the strategy. 

In practice, a skier can then get an upper-bound on the expected competitive ratio using only the prediction $\hat{\alpha}$ and the error $\epsilon$ in this prediction, both of which can come from a black-boxed machine learning algorithm. In the worst case, $\epsilon$ is bounded by $\max(\hat{\alpha},1-\hat{\alpha})$, and as displayed in Figure 3 a skier can, given a prediction $\hat{\alpha}$, determine the best expected competitive ratio possible (i.e. when the prediction is correct) and the worst expected competitive ratio possible (i.e. when the prediction is as wrong as possible). 

\begin{figure}
  \centering
  \includegraphics[width=0.8\textwidth]{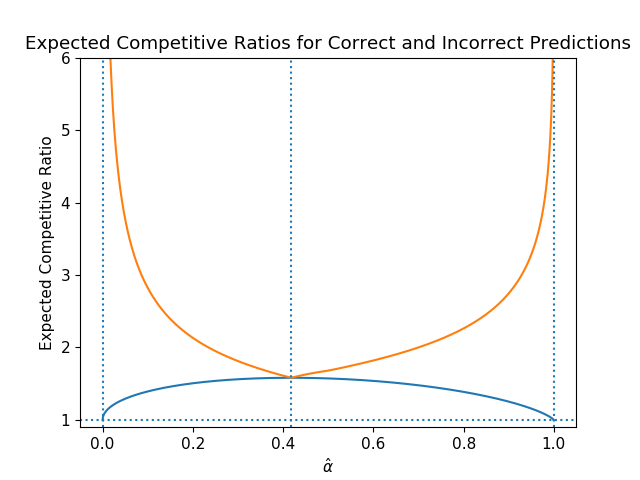}
  \caption{Competitive Ratio Range in the case of ML Errors}
\end{figure}

\section{Experiments}
We set $B=10$. We consider a case where a machine learning algorithm provides a prediction $\hat{\alpha}=0.15$ to the skier. Using Equation 5 in Section 2.1, we find numerically that $z^*(\hat{\alpha}) \approx 0.541$, and thus we set our cutoff for $p(x)$ at $Bz^*(\hat{\alpha}) = 5.41$. Based on sections 2.1 and 2.2, the skier will use a distribution 
$$ p(x) = \begin{cases} 
      \frac{e^{\frac{x}{10}}}{10 \left(e^{0.541}-1\right)} & \text{if } 0\leq x <5.41 \\
      0 & \text{otherwise}
   \end{cases}
$$
to pick the number of days to rent skis, and the adversary will use the distribution $q(y) = \frac{0.15  y e^{1-\frac{y}{10}}}{100(e-2)}$ when $y\in[0,B]$, $q(\infty) = 0.85$ to pick the number of ski days. We then take several draws of $x$ and $y$ from distributions $p(x)$ and $q(y)$ and calculate the competitive ratio in each case using Equation 2. 

For comparison, we also have the skier choose a cutoff using the optimal strategy without information, which is to pick $x$ using the distribution 
$$ p(x) = \begin{cases} 
      \frac{e^{\frac{x}{10}}}{10 \left(e-1\right)} & \text{if } 0\leq x <10 \\
      0 & \text{otherwise}
   \end{cases}
$$
Furthermore, to visualize the robustness of the algorithm we also have the skier choose $x$ using a distribution based on a faulty measurement $\hat{\alpha}=0.6$. In this case, the skier will use a distribution
$$ p(x) = \begin{cases} 
      \frac{e^{\frac{x}{10}}}{10 \left(e^{1.347}-1\right)} & \text{if } 0\leq x <13.47 \\
      0 & \text{otherwise}
   \end{cases}
$$
to pick $x$.

We then find a distribution of realized competitive ratios for each of the three algorithms used by drawing $x$ and $y$  10,000 times using each of these three distributions and calculating the competitive ratio for each trial. These results are summarized in Table 1. 

\begin{table}
  \caption{Empirical Performance of Ski Rental with Additional Information}
  \label{sample-table}
  \centering
  \begin{tabular}{lcl}
    Information Used &  Mean Competitive Ratio &  Theoretical Value\\
    \midrule
    Correct $\hat{\alpha} = 0.15$ & $1.446$ & $-0.85W_{-1}(-e^{-1/0.85}) \approx 1.45$    \\
    No $\hat{\alpha}$ provided    & $1.584$  & $\frac{e}{e-1} \approx 1.58$    \\
    Wrong $\hat{\alpha} = 0.60$     & $1.749$ & Upperbound : $1.45+(0.6-0.15)\Delta(0.15) \approx 1.95$ \\
    \bottomrule
  \end{tabular}
\end{table}

As expected, we find that correct information allows the skier to reduce the competitive ratio from around $1.58$ to around $1.45$ - this is close to the theoretical lower-bound on the best possible expected competitive ratio. Furthermore, we see that if wrong information is provided, the competitive ratio does suffer, but this loss of performance is well within the bounds established in Section 2.3.

\section{Conclusions}

In this paper we present a method to construct a probabilistic algorithm to optimally solve a variation of the Ski Rental problem in which the skier consults a machine learning algorithm to predict the probability that there will be at most $B$ ski days. This variant effectively utilizes the soft-threshold information that is the common output of many machine learning paradigms. We further establish how well this algorithm performs in the case of errors made by the machine learning algorithm, and provide performance guarantees in all cases. 

Our result is a novel way to incorporate information from machine learning algorithms into a classical problem of online optimization. The Ski Rental problem has various applications ranging from cloud server pricing to snoopy caching [4] - in many of these applications, predictive information is indeed available and thus our result provides a way to use this information while still preserving competitive guarantees.
\section*{References}

\small

[1] Purohit, Manish, Svitkina, Zoya \ \& Kumar, Ravi (2018) Improving Online Algorithms via ML Predictions. In {\it Advances in Neural Information Processing
  Systems 32}, pp.\ 9684-9693.

[2] Thodoris Lykouris and Sergei Vassilvitskii (2018) Competitive caching with machine learned advice.
In {\it ICML}, pp. \ 3302–3311.

[3] Anna R. Karlin, Mark S. Manasse, Lyle A. McGeoch, and Susan Owicki (1994) Competitive randomized algorithms for nonuniform problems. In {\it Algorithmica 11(6)}, pp. \ 542–571 .

[4] Anna R. Karlin, Mark S. Manasse, Larry Rudolph, and Daniel Dominic Sleator (1988) Competitive
snoopy caching. In {\it Algorithmica 3} , pp. \ 77–119.

[5] Andres Muñoz Medina and Sergei Vassilvitskii (2017) Revenue optimization with approximate bid
predictions. In {\it Advances in Neural Information Processing
  Systems 31} pp. \ 1856–1864.
  
  [6] Elias Koutsoupias and Christos Papadimitriou (1994) Beyond competitive analysis. In {\it Proc.
35th IEEE FOCS}, pp. 394-400.

[7] Hiroshi Fujiwara and Kazuo Iwama (2005) Average-case competitive analyses for ski rental
problems. In {\it Algorithmica}, vol. 42, no. 1, pp. 95–107.

[8] Yinfeng Xu and Weijun Xu (2004) Competitive algorithms for online leasing problem in
probabilistic environments,” in {\it Advances in Neural Networks (ISNN)} pp. 725–730.

[9] Tamar Basar and Geert J. Olsder (1999) Dynamic Noncooperative Game Theory. {\it SIAM
Series in Classics in Applied Mathematics}

\end{document}